\documentclass[letterpaper]{article} 
\usepackage[submission]{aaai25}  
\usepackage{times}  
\usepackage{helvet}  
\usepackage{courier}  
\usepackage[hyphens]{url}  
\usepackage{graphicx} 
\usepackage{natbib}  
\usepackage{caption} 
\usepackage{algorithm}
\usepackage{booktabs}       
\usepackage{amsfonts}       
\usepackage{nicefrac}       
\usepackage{xcolor}         
\usepackage{comment}
\usepackage{amsmath}
\usepackage{amssymb}
\usepackage{bm}
\usepackage{multirow}
\usepackage{pifont}
\usepackage{dsfont} 
\usepackage{mhchem}

\usepackage{algpseudocode}  
\usepackage{enumerate}
\usepackage[accsupp]{axessibility}

\usepackage{natbib}

\usepackage{ragged2e}
\usepackage{appendix}
\usepackage{booktabs}
\usepackage{color}
\usepackage{multicol}
\usepackage{caption}
\usepackage{marvosym}
\usepackage{newfloat}
\usepackage{listings}
\DeclareCaptionStyle{ruled}{labelfont=normalfont,labelsep=colon,strut=off} 
\floatstyle{ruled}
\newfloat{listing}{tb}{lst}{}
\floatname{listing}{Listing}
\newcommand{\method}{SHADE} 

\definecolor{symgreen}{RGB}{25, 142, 71}
\definecolor{symred}{RGB}{189, 31, 39}

\title{Human-aware 3D Scene Generation with Spatially-constrained Diffusion Models}
\author{
    Xiaolin, Hong\textsuperscript{\rm 1,3}\thanks{equal contribution}, 
    Hongwei, Yi\textsuperscript{\rm 2}\footnotemark[1], 
    Fazhi, He\textsuperscript{\rm 1}\footnote{corresponding author}
    Qiong, Cao\textsuperscript{\rm 3}\footnotemark[2]
}
\affiliations{
    \textsuperscript{\rm 1}School of Computer Science, Wuhan University\\
    \textsuperscript{\rm 2}{Hedra}\\
    \textsuperscript{\rm 3}JD Explore Academy\\
    \{xiaolin\_hong, fzhe\}@whu.edu.cn, hongwei@hedra.com, mathqiong2012@gmail.com\\
    https://hong-xl.github.io/SHADE
}

\usepackage{bibentry}

\begin{document}

\twocolumn[{
    \renewcommand\twocolumn[1][]{#1}
    \maketitle
    \centering
    \vspace{-4mm}
    \includegraphics[width=\textwidth]{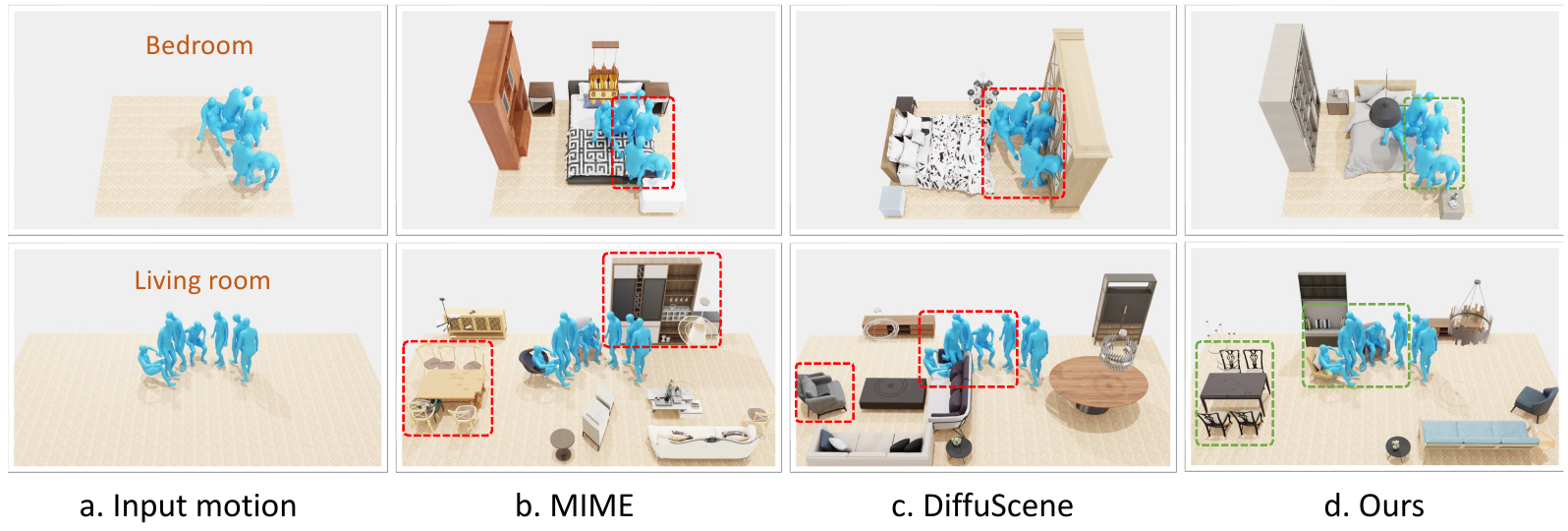}
    \captionof{figure}{
    Our method generates more plausible 3D scenes given input human motions and floor plans. It excels in two key aspects: (1) avoiding collisions between humans and objects, as well as between objects, a significant improvement over MIME \protect\cite{yi2022mime}, and (2) providing better support for human-object interactions compared to DiffuScene \protect\cite{tang2023diffuscene}.
    }
    \label{fig:teaser}
    \vspace{6pt}
}]

\begin{abstract}

Generating 3D scenes from human motion sequences supports numerous applications, including virtual reality and architectural design.
However, previous auto-regression-based human-aware 3D scene generation methods have struggled to accurately capture the joint distribution of multiple objects and input humans, often resulting in overlapping object generation in the same space. 
To address this limitation, we explore the potential of diffusion models that simultaneously consider all input humans and the floor plan to generate plausible 3D scenes.
Our approach not only satisfies all input human interactions but also adheres to spatial constraints with the floor plan.
Furthermore, we introduce two spatial collision guidance mechanisms: human-object collision avoidance and object-room boundary constraints. 
These mechanisms help avoid generating scenes that conflict with human motions while respecting layout constraints.
To enhance the diversity and accuracy of human-guided scene generation, we have developed an automated pipeline that improves the variety and plausibility of human-object interactions in the existing 3D FRONT HUMAN dataset.
Extensive experiments on both synthetic and real-world datasets demonstrate that our framework can generate more natural and plausible 3D scenes with precise human-scene interactions, while significantly reducing human-object collisions compared to previous state-of-the-art methods.
Our code and data will be made publicly available upon publication of this work.
\end{abstract}

%

\section{Introduction}
\label{sec:intro}

Creating diverse and realistic 3D environments inhabited by humans is essential for numerous applications, such as virtual reality (VR), interior design and training embodied artificial intelligence (AI) agents \cite{zhang2019survey}. 
This demand has driven researchers to explore diverse scene generation methods, propelling the rapid advancement of 3D scene synthesis \cite{luo2020end,Paschalidou2021NEURIPS,Liu2023CLIPLayoutSI}. 
Despite recent progress, there remain significant challenges in generating visually plausible scenes that adhere to various human motions.

There has been a recent surge in research dedicated to the problem of human-aware scene generation \cite{summon, yi2022mime}. A common approach in these studies is to learn an autoregressive model to sequentially place objects based on input humans and already generated objects. However, these methods often yield implausible scenes with object-object collisions. 
This limitation primarily arises from the inherent inability of autoregressive models to capture the joint distribution of multiple objects and multiple humans fully. 
Consequently, exploring a generative method that can effectively model and capture these complex distributions is crucial for generating realistic 3D scenes.

More recently, diffusion-based approaches for scene synthesis \cite{tang2023diffuscene, yang2024physcene} have shown an impressive ability to simplify the approximation of the joint distribution of objects, thereby generating the entire scenes at once and thus enhancing the realism of generated scenes. 
Meanwhile, many studies in image generation \cite{saharia2022palette,kawar2023imagic} and human motion synthesis \cite{huang2023diffusion,karunratanakul2023guided} have demonstrated that diffusion models can effectively incorporate inference guidance to meet user-defined goals.
Despite these advancements, there is still no standard solution for generating plausible 3D scenes that both support various human interactions and adhere to spatial constraints, such as avoiding motion collisions and respecting room boundaries. 

To tackle these challenges, we propose {{\method}}, a \textbf{S}patially-constrained \textbf{H}uman-\textbf{A}ware \textbf{D}iffusion based 3D \textbf{E}nvironments synthesis. 
As shown in Figure \ref{fig:teaser}, our method can generate plausible scene layouts that avoid collisions with humans and between objects, while supporting various human activities such as sitting and lying. 
Our key insight lies in innovative harnessing diffusion models to simultaneously input all humans and the floor plan to generate holistic object configurations. 
Specifically, we input the contact bounding boxes and free space extracted from input human motions and the floor plan, following \cite{yi2022mime}. 
We then learn a diffusion model to capture the joint distribution of objects, enabling the simultaneous generation of object placements and understanding the relationships between their attributes. 
To further enhance the plausibility of generated scenes, we design two spatial collision guidance functions:
1) motions collision avoidance that calculates the collision ratio between objects and moving humans to prevent their implausible penetrations, while 2) boundary constraint that penalizes the distance by which objects extend beyond the floor plan, ensuring that object placement respects room boundaries.
During inference, we combine two guidance functions with an object-object collision function \cite{yang2024physcene}. This allows our diffusion model to generate collision-free scenes that respect human movements, room boundaries, and prevent object overlap.

Beyond model design, we enhance human-aware scene generation by addressing key data challenges in datasets like 3D FRONT HUMAN \cite{yi2022mime}. 
Specifically, we tackle two main issues:
1) incorrect penetrations in human-object interactions hinder accurate spatial relationship modeling,
and 2) limited interaction diversity restricts model generalization. 
Our approach includes automated techniques for adjusting translations to correct penetrations and augmenting categories and orientations to diversify human-object interactions.
We use our calibrated dataset to train {{\method}} and evaluate it on synthetic and real datasets. Our results, both quantitative and qualitative, show that our method generates more plausible 3D scenes with realistic human-scene interactions and effectively reduces human-object collisions. These findings underscore the superiority of {{\method}} compared to previous state-of-the-art methods.

Our main contributions are:
(i) We propose a human-aware diffusion-based model for generating realistic 3D scene layouts with plausible human-object interactions in a single step.
(ii) We design two spatial collision inference strategies, motion collision avoidance and boundary constraint guidance, to enhance scene fidelity by preventing collisions with human motions and respecting room layouts.
(iii) We devise an automated calibration pipeline to enhance accuracy and diversity in human-object interactions within the original dataset, thereby improving the generative quality and diversity of human-aware scene synthesis.

\section{Related Works}
\noindent\textbf{Human-agnostic Scene Synthesis.}
The objective of human-agnostic scene generative methods is to generate plausible and diverse scene layouts without taking into account human activities. Early works explore procedural modeling with grammars \cite{muller2006procedural} to place objects in the scenes, heavily relying on manually designed rules and producing scenes with limited diversity.
Subsequently, graph-based approaches \cite{Zhou_2019_ICCV, luo2020end, 2023_gao_sceneHGN} have been extensively developed to represent 3D scenes as scene graphs and capture the underlying structure of scenes using graph neural networks. 
Unlike these studies, recent autoregressive generation methods \cite{Ritchie_2019_CVPR, wang2020sceneformer, Paschalidou2021NEURIPS, Liu2023CLIPLayoutSI} regard 3D scenes as object sequences and learn autoregressive models to sequentially predict the next object conditioned on already generated objects. 
 However, these autoregressive methods fail to accurately capture the distribution of the entire object sequence, which often results in generating overlapping objects in the same place. More recently, some diffusion-based approaches \cite{tang2023diffuscene, yang2024physcene} have been proposed to ease the approximation of the joint distribution of multiple objects, benefiting the generation of high-fidelity scenes.  
Inspired by these works, we develop a spatially-constrained diffusion method, {\method}. 
This framework learns a scene diffusion model to capture the joint distribution of objects conditioned on both human motions and the floor plan, and incorporates spatial-constrained guidance to improve scene plausibility. 

\begin{figure*}[t!]
\centering
\vspace{-5mm}
\includegraphics[width=17cm]{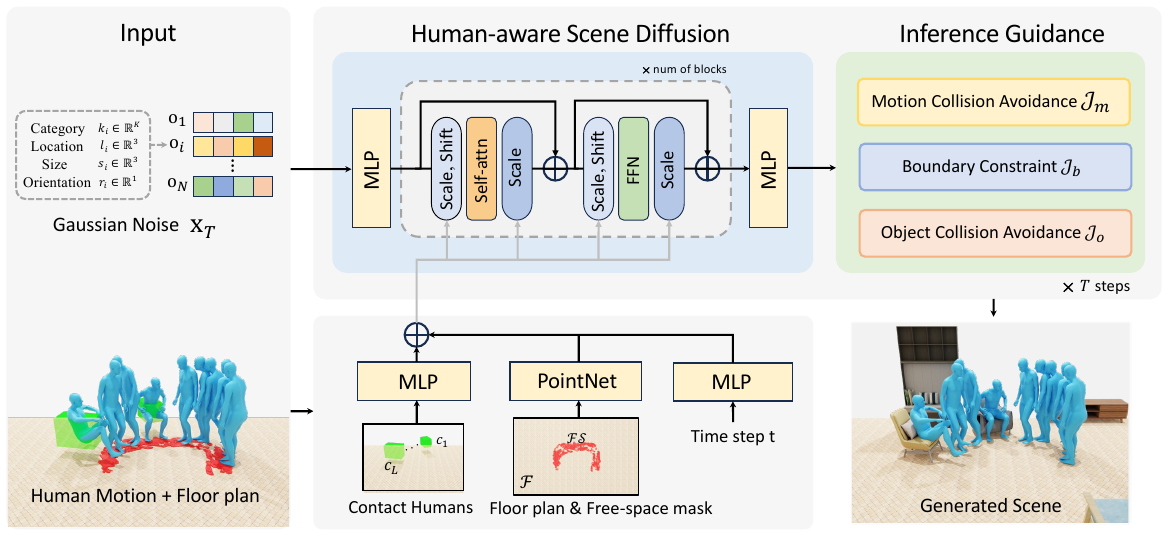}
\caption{
\textbf{Overview of our method.} {{\method}} learns a diffusion model to gradually clean the noisy scene $\mathbf{x}_T$ by simultaneously considering the contact bounding boxes, free-space mask, floor plan, and time step. During inference, {\method} applies three spatial collision guidance functions to ensure the generation of plausible scenes that avoid conflicts with human motions, room boundaries, as well as prevent object overlap.
}
\vspace{-4mm}
\label{overview}
\end{figure*}

\noindent\textbf{Human-aware Scene Synthesis.}
This branch of scene synthesis focuses on producing plausible scenes in which the input human motions can naturally take place. To this end, Pose2Room \cite{nie2022pose2room} proposes a pose-conditioned generative model to predict object configurations from human pose trajectories. 
However, Pose2Room can only predict contact objects rather than an entire scene. In contrast, SUMMON \cite{summon} learns a ContactFormer to generate affordable objects that contact with humans and employs an autoregressive model to complete the scene. 
Similarly, MIME \cite{yi2022mime} learns an autoregressive model to sequentially predict object placements based on the contact information and free space extracted from human motions and floor plan. 
Unlike these approaches, {\method} is a diffusion-based, non-autoregressive method that inherently explores the relationships between all object attributes to generate more plausible furniture layouts. 


\section{Method}

The overview of {\method} is presented in Fig. \ref{overview}. 
Specifically, our method takes in the contact bounding boxes and free space extracted from given human motions and the floor plan, and extracts their embeddings with respective encoders.  
Conditioned on these embeddings, we learn a scene diffusion model to capture the joint distribution of multiple objects. This enables us to generate plausible object configurations by exploring the relationships between their attributes. 
At inference time, we apply three spatial collision guidance functions to guide the diffusion model in generating plausible 3D scenes that avoid conflicts with human motions and respect layout constraints.
%
In the following, we outline the problem formulation and detail the human-aware scene diffusion model and inference guidance.

\subsection{Problem Formulation}
\label{setup}
Given input human motions $\mathcal{H}$ and an empty floor plan $\mathcal{F}$, our goal is to generate a 3D scene $\mathcal{S}$ that can support various human interactions and movements. 
Following \cite{yi2022mime,tang2023diffuscene}, the scene $\mathcal{S}$ is represented as an unordered set of $N$ objects, denoted as $\mathbf{x}=\{o_i\}^{N}_{i=1}$. Each object $o_i=\{k_i, \ell_i, s_i, r_i\}$ consists of a semantic label $k_i \in \mathbb{R}^{K}$ out of $K$ categories, location $\ell_i \in \mathbb{R}^{3}$, size $s_i \in \mathbb{R}^{3}$ and orientation $r_i \in \mathbb{R}^{1}$. 
These objects $\mathcal{O}$ can be categorized into two kinds: 1) contact objects $\mathcal{Q}=\{q\}^{L}_{i=1}$, which interact with humans, and 2) non-contact objects $\bar{\mathcal{Q}}=\{\bar{q}\}^{N-L}_{i=1}$, which do not.
To condition the 3D scene generation with human motions,  both contact humans and free-space humans are extracted from the input motions, following \cite{yi2022mime}. Specifically, contact humans indicate the location and category of contact objects, represented as the collection of contact boxes $\mathcal{C}=\{c_i\}^{L}_{i=1}$. Free-space humans define the walkable area of a room, specifying regions where objects cannot be placed. This information is represented as a binary free-space mask $\mathcal{FS}$ by projecting all foot contact points on the floor plan. 
Formally, conditioned on the floor plan $\mathcal{F}$, the free-space mask $\mathcal{FS}$ and all contact humans $\mathcal{C}$, we learn a generative model to predict the contact objects $\mathcal{Q}$ and the non-contact objects $\bar{\mathcal{Q}}$, such that they can support all human interactions while adhering to the constraints imposed by human motions $\mathcal{H}$ and the floor plan $\mathcal{F}$.

\subsection{Human-aware Scene Synthesis}\label{method}
\label{sec:diffusion}
The diffusion model provides a robust framework for scene generation by learning the joint distribution of multiple objects, which is essential for creating plausible scenes without overlapping objects. Next, we detail how we incorporate the diffusion model with all input humans and the floor plan to generate realistic 3D scenes.

Denote $\mathbf{x}_0\sim q(\mathbf{x}_0)$ as a clean scene sampled from the training data, we gradually add Gaussian noise to $\mathbf{x_0}$ with a forward diffusion process $q(\mathbf{x}_{t+1}|\mathbf{x}_t)$ of length $T$. After $T$ diffusion steps, it approximates Gaussian noise $\mathbf{x}_T\sim \mathcal{N}(\mathbf{0}, \mathbf{I})$. 
To generate scenes conditioned on human motions and floor plan, our diffusion model learns a reverse denoising process $p_\theta(\mathbf{x}_{t-1}|\mathbf{x}_{t}, C)$ to convert $\mathbf{x}_T$ back to $\mathbf{x}_0$:
\begin{equation}
p_\theta(\mathbf{x}_{t-1}|\mathbf{x}_{t}, C)=\mathcal{N}(\mathbf{x}_{t-1};\mu_\theta(\mathbf{x}_t, t, C), (1-\alpha_{t})\mathbf{I}), \label{eq1}
\end{equation} 
where $\theta$ are the model parameters, $\alpha_{t}$ depends on a pre-defined variance schedule, $C$ denotes the conditioning signals. 
As illustrated in Fig. \ref{overview}, we select $C=\{ \mathcal{C}, \mathcal{F}, \mathcal{FS}\}$ as the conditioning signals, enabling the diffusion model to generate 3D scenes by leveraging the information from all contact humans $\mathcal{C}$, the 2D floor plan $\mathcal{F}$, and the 2D free-space mask $\mathcal{FS}$.
According to $\epsilon-$prediction \cite{ho2020denoising}, $\mu_\theta(\mathbf{x}_t, t, C)$ can be re-parameterized as:
\begin{equation}
\mu_\theta(\mathbf{x}_t, t, C)=\frac{1}{\sqrt{\alpha_t}}\left(\mathbf{x}_t-\frac{1-\alpha_t}{\sqrt{1-\bar{\alpha}_t}}{\bm{\epsilon}_\theta(\mathbf{x}_t, t, C)}\right),\label{eq2}
\end{equation}
where $\bar{\alpha}_t=\prod_{i=1}^t \alpha_i$, $\bm{\epsilon}_\theta(\mathbf{x}_t, t, C)$ is a denoising network that predicts the noise applied to $\mathbf{x}_0$ from a noisy scene $\mathbf{x}_t$ and the  conditions $C$. Finally, we can reconstruct a clean scene $\mathbf{x_0}$ as follows:
\begin{equation}
p_\theta(\mathbf{x}_{0}|C)=p(\mathbf{x}_T)\prod_{t=1}^{T}p_\theta(\mathbf{x}_{t-1}|\mathbf{x}_{t}, C),
\end{equation}
where $p_\theta(\mathbf{x}_{0}|C)$ denotes the probability of scene $\mathbf{x}_{0}$ conditioned on $C$. Mathematically, we can maximize the conditional probability
$p_\theta(\mathbf{x}_{0}|C)$ by training the denoising network $\bm{\epsilon}_\theta(\mathbf{x}_t, t, C)$ with a simplified objective \cite{ho2020denoising}:
\begin{equation}
\begin{aligned}
\mathcal{L}_{sim}
&=\mathbb{E}_{\bm{\epsilon}, t, \mathbf{x}_0}\left[\left\|\bm{\epsilon}-\bm{\epsilon}_\theta(\sqrt{\bar{\alpha}_t}\mathbf{x}_0+\sqrt{1-\bar{\alpha}_t}\bm{\epsilon}, t, C)\right\|^2\right].
\end{aligned}
\end{equation}

During scene generation, our diffusion model starts from a Gaussian noise $\mathbf{x}_T$ and gradually predicts a cleaner scene with Eq. \ref{eq1} and \ref{eq2}. 
After $T$ steps, we generate a plausible scene $\tilde{\mathbf{x}}_0$ that can interact with various human motion sequences, such as touching, sitting, and lying.   

\subsection{Inference Guidance for Spatial Constraints}\label{guidance}

To further enhance the plausibility of 3D scenes during the diffusion sampling, we introduce two spatial collision guidance functions: 1) the motion collision avoidance function pushes objects back when they collide with human motions, and 2) the boundary constraint function repositions objects that are outside the floor plan.

\noindent\textbf{Motion Collision Avoidance.}   
We estimate motion collision scores using the 3D bounding boxes of predicted objects $\{\hat{o}_i\}^{N}_{i=1}$ and 2D free-space mask $\mathcal{FS}$ induced by human movements. 
Specifically, we project the predicted objects onto the floor plan and calculate the ratio of the free space mask collided by the 2D projections of these objects:
\begin{equation}
\mathcal{J}_m=\left(\sum_{i=1}^{N}\sum_{p\in \hat{o}_i}^{} \mathcal{FS}(p)\right) / \sum_{p\in \mathcal{FS}}\mathcal{FS}(p),
\end{equation}
where $p$ denotes each pixel on the 2D free-space mask image at $256^2$ resolution.

\noindent\textbf{Boundary Constraint.} 
Besides satisfying motion collision constraints, the generated objects must also meet boundary constraints, meaning all objects should be located within the given floor plan. 
To this end, we introduce an additional boundary constraint function to penalize objects that exceed the boundaries. Similar to motion collision avoidance, we compute the collision ratio between the region outside the floor plan and the 2D projections of the predicted objects:
\begin{equation}
\mathcal{J}_b=\left(\sum_{i=1}^{N}\sum_{p\in \hat{o}_i}^{} \bar{\mathcal{F}}(p)\right) / \sum_{p\in \bar{\mathcal{F}}}\bar{\mathcal{F}}(p),
\end{equation}
where $\bar{\mathcal{F}}=1-\mathcal{F}$ denotes the region outside the floor plan.

\noindent\textbf{Overall Inference Guidance.} To generate more plausible scenes that respect spatial constraints from human motions and room layout, we integrate motion collision avoidance $\mathcal{J}_m$ and boundary constraint $\mathcal{J}_b$ into the inference guidance of our scene diffusion model. Inspired by \cite{yang2024physcene}, we also employ the object collision avoidance $\mathcal{J}_o$ to penalize the collision between objects. Finally, our overall inference guidance is formulated as $\mathcal{J}=\mathcal{J}_m+\mathcal{J}_b+\mathcal{J}_o$, where 
$\mathcal{J}_o=\sum_{i,j,i\neq j} \mathbf{IoU}({o}_i, o_j)$ measures the objects collision score by calculating the 3D IoU of each object pair. 
During generative process, we first estimate the $\hat{\mathbf{x}}_0$ from the predicted noise via 
$\hat{\mathbf{x}}_0=\frac{1}{\bar{\alpha}_t}\left(\mathbf{x}_t-\sqrt{1-\bar{\alpha}_t}\bm{\epsilon}_\theta(\mathbf{x}_t, t, C)\right),$
then calculate the gradient of $\mathcal{J}$ to perturb the predicted $\hat{\mathbf{x}}_0$ at the denoising step $t$ via
$\tilde{\mathbf{x}}_0 = \hat{\mathbf{x}}_0 - \gamma\nabla_{\mathbf{x}_t}\mathcal{J}(\hat{\mathbf{x}}_0, \mathcal{FS}, \mathcal{F})$,
where $\gamma$ controls the guidance strength. Next, the predicted mean $\mathbf{\mu}_\theta$ is computed with the updated clean scene $\tilde{\mathbf{x}}_0$ as \cite{ho2022video}. 
Please refer to Sup. Mat. for more details.

\section{Dataset Calibration of 3D FRONT HUMAN}

\begin{figure*}[h]
\centering
\vspace{-5mm}
\includegraphics[width=17cm]{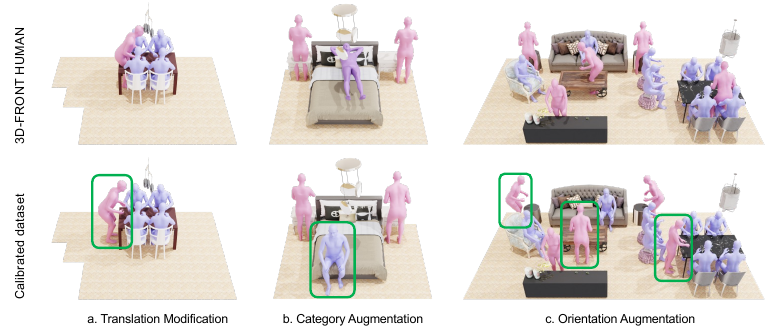}
\caption{Comparison between 3D-FRONT HUMAN \cite{yi2022mime} and our calibrated dataset. We correct human-object penetrations through translation modification to improve spatial accuracy. Additionally, we apply category and orientation augmentation to enhance the diversity in interactions.}
\vspace{-4mm}
\label{exp:compare_dataset}
\end{figure*}

In this section, we introduce the calibration pipeline designed to tackle two major data challenges in the 3D FRONT HUMAN dataset \cite{yi2022mime}. 
Our pipeline begins with translation modification to correct penetrations in human-object interactions, thereby improving spatial accuracy. Next, it employs category and orientation augmentation to increase the diversity of interactions, which enhances model generalization. These techniques are explained in greater detail below.

\begin{table}
\renewcommand{\arraystretch}{1.6}
\resizebox{\linewidth}{!}{
\fontsize{16}{14}\selectfont
\begin{tabular}{cccc}
\toprule
{Room Type} & {Dataset} & {$\mathbf{E_{pen}}$ $\downarrow$} & {3D IoU $\uparrow$} \\ 
\midrule
\multirow{2}{*}{Bedroom} & 3D-FRONT HUMAN \cite{yi2022mime} & 226.97 & 0.75 \\
& Our calibrated dataset & \textbf{16.01} & \textbf{0.94} \\ 
\midrule
\multirow{2}{*}{Living} & 3D-FRONT HUMAN \cite{yi2022mime} & 125.49 & 0.77 \\
& Our calibrated dataset & \textbf{19.50} & \textbf{0.94}  \\ 
\midrule
\multirow{2}{*}{Dining} & 3D-FRONT HUMAN \cite{yi2022mime} & {133.34} & 0.78 \\
& Our calibrated dataset & \textbf{20.72} & \textbf{0.95} \\ 
\bottomrule
\end{tabular}
}
\caption{Quantitative comparisons between the 3D-FRONT HUMAN \cite{yi2022mime} and our calibrated dataset on human-scene interaction metrics.}
\vspace{-4mm}
\label{tab:dataset}
\end{table}

\noindent\textbf{Translation Modification.} 
To address incorrect penetrations in human-object interactions, we adjust the translation parameters of humans to avoid implausible contact with objects. Specifically, we modify these parameters within a range of $[-2m, 2m]$ centered on the contact object. We then evaluate the plausibility of the interactions using two indicators: the 3D IoU score \cite{zhou2019iou} between the bounding boxes of the human and object, and the human-scene interpenetration error $\mathbf{{E}_{pen}}$ \cite{shen2023learning}:
\begin{equation}
\mathbf{{E}_{pen}}=\sum_{i=1}^V \mathds{1}_{x<0}\left[sdf\left(v_i, \mathcal{S}\right)\right] \cdot\left|s d f\left(v_i, \mathcal{S}\right)\right|,
\end{equation}
where $V$ is the number of human mesh vertices, $sdf\left(v_i, \mathcal{S}\right)$ is the signed distance from vertex $v_i$ to scene $\mathcal{S}$, and $\mathds{1}_{x<0}[\cdot]$ is an indicator function that returns 1 when the condition is met, and 0 otherwise. 
The translation modification is considered complete when $\mathbf{{E}_{pen}}$ falls below the safe bound $\sigma_1$ and the 3D IoU is higher than $\sigma_2$. Empirically, we set $\sigma_1=20$, $\sigma_2=90\%$ for sitting and lying humans, and $\sigma_2=50\%$ for touching humans.


\noindent\textbf{Category Augmentation.} 
After refining the implausible human-object interpenetration,
we perform category augmentation to enhance the diversity of human-object interactions.
The motivation behind category augmentation is that people interact with objects in various ways in real human activities, such as lying on a bed or sitting on it.
To capture these behaviors, we augment the dataset by randomly replacing half of the contact humans with alternate modes of interaction, thereby fabricating a refreshed set of human-object interactions. 
As shown in Fig. \ref{exp:compare_dataset} b, we transform a scene from ``lying on the bed" to ``sitting on the bed". 

\begin{figure*}[h]
\centering
\vspace{-5mm}
\includegraphics[width=\textwidth]{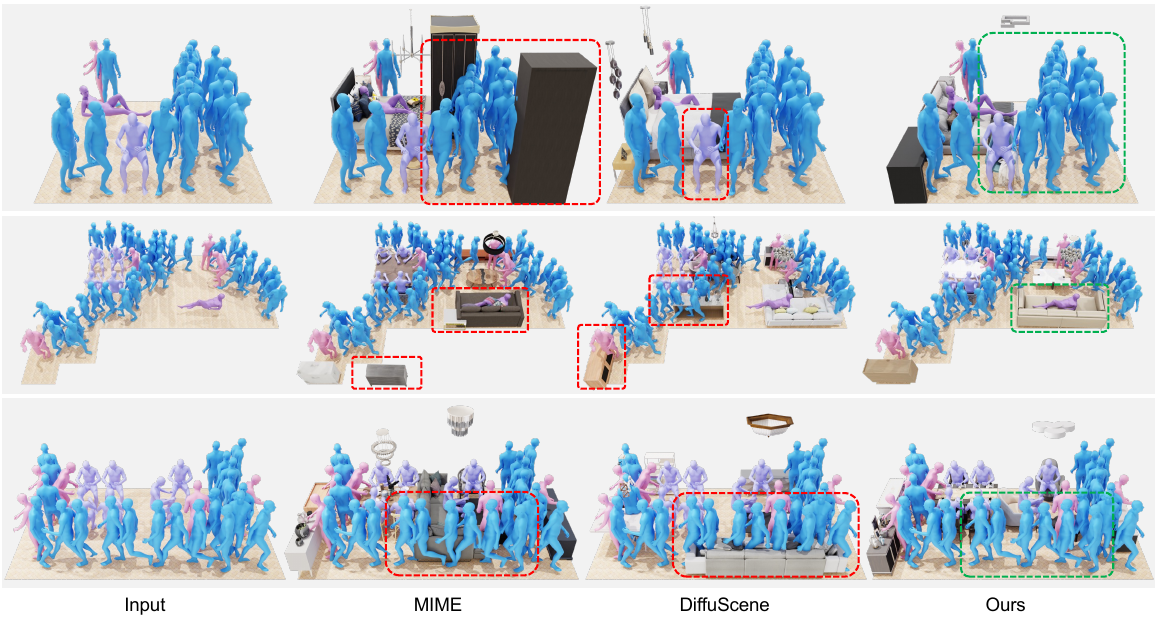}
\caption{\textbf{Qualitative comparison on the test split in calibrated 3D FRONT HUMAN}. 
Compared with existing state-of-the-art methods MIME and DiffuScene, our method generates more plausible scenes that avoid conflict with free-space humans and room boundaries, and present fewer overlapping objects. Each row represent an example input.
}
\label{exp:compare}
\end{figure*}

\begin{table*}[h!]
\centering
\renewcommand{\arraystretch}{1.0}
\begin{tabular}{cccccccc}
\toprule
{Room Type} & {Method} & {3D IoU $\uparrow$} & {$\mathbf{Col_{mot}}$ $\downarrow$} & {$\mathbf{R_{out}}$ $\downarrow$} & {$\mathbf{Col_{obj}}$ $\downarrow$} & {FID $\downarrow$} & {CKL $\downarrow$}  \\ 
\midrule
\multirow{3}{*}{Bedroom} & MIME \cite{yi2022mime} & \underline{0.905} & \underline{0.068} & \underline{0.003} & {0.020} & \underline{38.06} & {0.033}  \\
& DiffuScene \cite{tang2023diffuscene} & 0.550 & {0.150} & {0.004} & \underline{0.015} & 40.90 & \underline{0.018} \\
& {\method} (Ours) & \textbf{0.925} & \textbf{0.027} & \textbf{0.003} & \textbf{0.001} & \textbf{37.62} & \textbf{0.007} \\ 
\midrule
\multirow{3}{*}{Living} & MIME \cite{yi2022mime} & \underline{0.899} & \underline{0.032} & \underline{0.002} & \underline{0.062} & \underline{38.75} & \textbf{0.012} \\
& DiffuScene \cite{tang2023diffuscene} & 0.301 & {0.165} & {0.003} & 0.063 & 40.89 & 0.021  \\
& {\method} (Ours) & \textbf{0.939} & \textbf{0.029} & \textbf{0.001} & \textbf{0.038} & \textbf{37.85} & \underline{0.014}   \\ 
\midrule
\multirow{3}{*}{Dining} & MIME \cite{yi2022mime} & \underline{0.924} & \underline{0.031} & \underline{0.002} & \underline{0.059} & \textbf{38.02} & {0.021} \\
& DiffuScene \cite{tang2023diffuscene} & 0.343 & {0.157} & \underline{0.002} & 0.077 & 41.98 & \underline{0.012} \\
& {\method} (Ours) & \textbf{0.943} & \textbf{0.028} & \textbf{0.001} & \textbf{0.049} & \underline{39.63} & \textbf{0.009}   \\ 
\bottomrule
\end{tabular}
\caption{Quantitative comparison on the test split of the Calibrated 3D FRONT HUMAN dataset.
We compare {\method} with MIME and DiffuScene on human-scene interaction score 3D IoU, scene plausibility metrics $\mathbf{Col_{mot}}$, $\mathbf{R_{out}}$, $\mathbf{Col_{obj}}$, and standard perceptual quality scores FID, CKL. 
The best scores are highlighted in bold, and the second-best scores are underlined.}
\vspace{-4mm}
\label{res}
\end{table*}

\begin{table}[h!]
\centering
\renewcommand{\arraystretch}{1.2}
\resizebox{\linewidth}{!}{
\begin{tabular}{ccccc}
\toprule
 {Method} & {3D IoU $\uparrow$} & {$\mathbf{Col_{mot}}$ $\downarrow$} & {$\mathbf{R_{out}}$ $\downarrow$} & {$\mathbf{Col_{obj}}$ $\downarrow$} \\ 
\midrule
MIME \cite{yi2022mime} & {0.882} & 0.176 & {0.003} & 0.109 \\
DiffuScene \cite{tang2023diffuscene} &0.036  &0.152  &\textbf{0.002}  &0.073  \\
{\method} (Ours) & \textbf{0.904} & \textbf{0.135} & {0.003} & \textbf{0.014} \\ 
\bottomrule
\end{tabular}
}
\caption{Quantitative comparison on the PROXD qualitative dataset \cite{hassan2019resolving}.}
\vspace{-6mm}
\label{res_prox}
\end{table}

\noindent\textbf{Orientation Augmentation.} 
Considering that people might interact with objects from various angles, we also introduce orientation augmentation to enrich the dataset. 
To achieve this, we firstly add random noise to orientation parameters of contact humans. Next, we adjust the translation parameters of contact humans until their interactions with objects meet the same criterion proposed in the translation modification.
This ensures that contact humans can interact with objects in a varied and realistic manner.

To verify the effectiveness of our calibrated pipeline, we report $\mathbf{E_{pen}}$ and 3D IoU score for 3D-FRONT HUMAN and our calibrated dataset in Tab. \ref{tab:dataset}. It shows that the calibration pipeline effectively reduces human-scene interpenetration errors and improves human-object interactions of 3D-FRONT HUMAN across all room types. 
The qualitative comparisons presented in Figure \ref{exp:compare_dataset} also demonstrate that our calibrated dataset provides more plausible and diverse human-object interactions.

\section{Experiments} \label{exp}
\noindent\textbf{Dataset.} We conduct experiments on the calibrated 3D FRONT HUMAN dataset, which contains a total of 5,689 bedrooms, 2,987 living rooms, and 2,549 dining rooms. Following \cite{yi2022mime}, we use 21 object categories for the bedrooms, and 24 for the living rooms and dining rooms. For each kind of room, we split the data into 80\% for training, 10\% for validation, and 10\% for testing. We train and validate our model on the training and validation sets respectively, and evaluate it on the test set for each room type.

\noindent\textbf{Baselines.} We compare our method with MIME \cite{yi2022mime} and DiffuScene \cite{tang2023diffuscene} using their official implementations. 
MIME is an autoregressive method for human-aware scene generation, while DiffuScene is a diffusion-based method that learns 3D scene distributions without conditioning on the floor plan. For a more fair comparison, we adapt DiffuScene on the 2D-floor plan with the free-space mask to enhance its perception of human motions. All methods are trained on our calibrated datasets.

\noindent\textbf{Evaluation Metrics.}
Following \cite{yi2022mime}, we evaluate all methods on the plausibility of human-object interaction and scene realism. 
To measure the plausibility of human-object interaction, we estimate the 3D IoU score by calculating the intersection ratio between input contact bounding boxes and generated objects. 
We also employ $\mathbf{Col_{mot}}$ \cite{yi2022mime} to evaluate collision between generated objects and free-space humans, and $\mathbf{Col_{obj}}$ \cite{yang2024physcene} to represent the collision rate between objects and objects. 
To measure the violation of the room layout, we compute the collision rate between generated objects and the areas outside the floor plan, denoted as $\mathbf{R_{out}}$. 
For realism, we calculate the \text{Fréchet} inception distance (FID) score and the category KL divergence (CKL) \cite{Paschalidou2021NEURIPS} between generated and real scenes. 
All evaluation experiments are conducted on the test split of the calibrated datasets.

\noindent\textbf{Network Architecture.} 
As depicted in Fig. \ref{overview}, we implement {\method} with a transformer-based architecture composed of four DiT blocks with adaLN-Zero \cite{2022_scale}. 
As input, all object parameters are encoded by a fully connected network, while the time step $t$ is encoded using a positional embedding \cite{tancik2020fourier}. 
To make the scene generation aware of human motion sequences and room layout, contact humans, the floor plan $\mathcal{F}$ and the free-space mask $\mathcal{FS}$ are encoded by a fully connected network and PointNet \cite{qi2017pointnet}, respectively. These embeddings adjust the scale and shift parameters of each adaLN-Zero block. 
See more details in Sup. Mat. 

\noindent\textbf{Implementations.}
We train our human-aware scene diffusion model with the default settings in \cite{NEURIPS2020_ddpm}, and reduce the number of diffusion steps to $T$=200 from 1000 for speeding up the sampling process.
See more details in Sup. Mat.  
At inference time, we use the DDPM sampler \cite{NEURIPS2020_ddpm} to obtain the object properties. 
Subsequently, we retrieve the closest CAD model from the 3D-FUTURE dataset \cite{future20213d} based on the class label and bounding box size of the generated objects.

\begin{table}[]
\centering
\renewcommand{\arraystretch}{1.4}
\resizebox{\linewidth}{!}{
\begin{tabular}{ccc|ccc} \hline
Motion Collision  &Room Boundary &Object Collision &{$\mathbf{Col_{mot}}$($\downarrow$)} &$\mathbf{R_{out}}$($\downarrow$)  &$\mathbf{Col_{obj}}$($\downarrow$)  \\ \hline\hline
\textcolor{symred}{\ding{55}}  &\textcolor{symred}{\ding{55}}  &\textcolor{symred}{\ding{55}}  &0.043   &\underline{0.002}      &0.013    \\ 
\textcolor{symgreen}{\ding{51}}  &\textcolor{symred}{\ding{55}}  &\textcolor{symred}{\ding{55}}     &\underline{0.034}    &\underline{0.002}    &0.013  \\ 
\textcolor{symred}{\ding{55}} &\textcolor{symgreen}{\ding{51}} &\textcolor{symred}{\ding{55}} &0.045   &\textbf{0.001}  &0.014   \\
\textcolor{symred}{\ding{55}} &\textcolor{symred}{\ding{55}} &\textcolor{symgreen}{\ding{51}} &0.036   &0.003   &\underline{0.001}   \\ 
\textcolor{symgreen}{\ding{51}} &\textcolor{symgreen}{\ding{51}} &\textcolor{symgreen}{\ding{51}} &\textbf{0.027}   &{0.003}   &\textbf{0.001} \\ \hline
\end{tabular}8
}
\caption{Ablation study on different spatial collision guidance functions. The best scores are highlighted in bold, and the second best scores are underlined.}
\vspace{-5mm}
\label{tab:ablation}
\end{table}


\subsection{Human-aware Scene Synthesis}
Figure \ref{exp:compare} exhibits the generation ability of our method and baselines for different room types. 
Since DiffuScene only considers floor plan and free-space information, it fails to generate appropriate objects to interact with contact humans. Additionally, DiffuScene generates objects colliding with free-space humans. 
Both MIME and our method can generate reasonable objects to support various contact humans, such as placing a bed under a sitting human or a sofa under a lying human.
However, MIME still generates objects in free-space humans or outside the floor plan (see the second column of Fig. \ref{exp:compare}). In contrast,  our method generates more plausible scenes that can avoid colliding with free-space humans and respect room boundaries, as well as prevent object overlap.
%
These observations are validated by the quantitative comparisons presented in Table \ref{res}. 
Our method outperforms the baselines in 3D IoU scores and significantly reduces motion collisions ($\mathbf{Col_{mot}}$), object collisions ($\mathbf{Col_{obj}}$) and room layout violations ($\mathbf{R_{out}}$) across all room types.
Additionally, our method achieves the best FID scores in the Bedroom and Living room settings, and the best CKL scores
in the Bedroom and Dining room settings, indicating that our method generates more realistic scenes.
Overall, these results demonstrate the effectiveness of {\method}.

Following MIME \cite{yi2022mime}, we test {\method} on a real dataset of human motion to evaluate the generalization of our method. We use 5 input motions from the PROXD \cite{hassan2019resolving} dataset and generate 10 scenes for each motion sequence. The quantitative results of our method and the baselines are reported in Table \ref{res_prox}.  
All methods are not fine-tuned on the PROXD dataset. 
For human-object interaction plausibility, our method achieves a 3D IoU score of 0.881, outperforming MIME and DiffuScene.
In terms of scene plausibility, our method surpasses the baseline methods on the motion collision metric $\mathbf{Col_{mot}}$ and the object collision metric $\mathbf{Col_{obj}}$, indicating that our method generates more realistic scenes with fewer human-object and object-object collisions. 
Figure \ref{fig:teaser} presents the qualitative visualization of all models’ generations.

\begin{figure}[h]
\centering
\vspace{-1mm}
\includegraphics[width=\columnwidth]{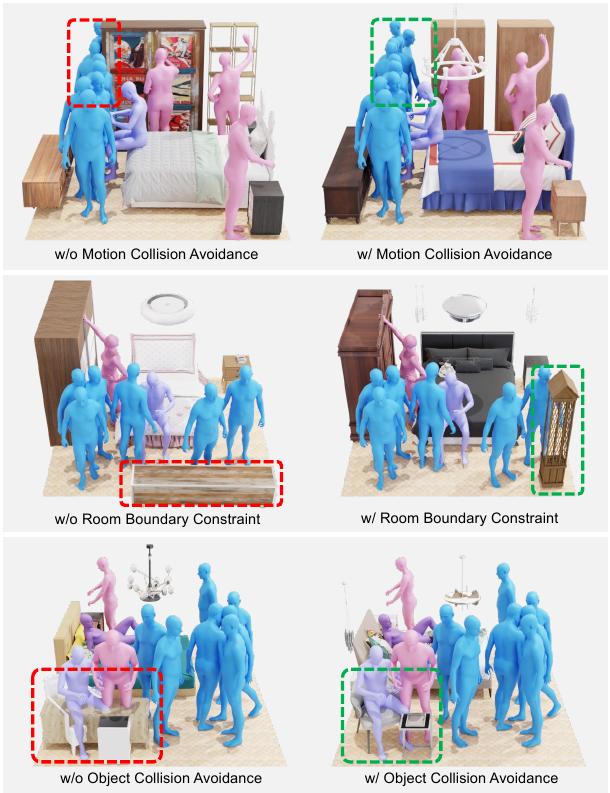}
\caption{
Ablation on spatial collision guidance functions. The left column shows scenes generated without guidance, with red boxes indicating constraint violations. The right column shows scenes with guidance, where green boxes highlight the improvements.
}
\label{ablation1}
\vspace{-3mm}
\end{figure}

\subsection{Ablation Study on Inference Guidance}
We investigate the impact of each spatial collision guidance function on the bedroom and present the results in Tab. \ref{tab:ablation}. 
Compared to scene generation without any guidance function, incorporating the motion collision avoidance function reduces the $\mathbf{Col_{mot}}$ metric to 0.034 from 0.043, demonstrating its effectiveness.
Similar conclusions can be drawn from the results of the other two spatial collision guidance functions. 
It is noteworthy that these collision-based guidance functions can negatively affect each other. For example, while the room boundary constraint function improves the $\mathbf{R_{out}}$ metric, it can degrade the other metrics. This is reasonable because room boundary guidance pushes objects within the floor plan, potentially leading to increased collisions with free-space humans ($\mathbf{Col_{mot}}$) and other objects ($\mathbf{Col_{obj}}$).  
To balance these effects, we integrate all spatial collision functions, thereby achieving better overall performance in scene plausibility.
Fig. \ref{ablation1} provides a qualitative visualization of the effect of each spatial collision guidance function. 
The improvements shown in Fig. \ref{ablation1} confirm that our collision-based guidance functions can significantly enhance the plausibility of 3D scenes.

\section{Conclusions}

We introduced {\method}, a spatially-constrained diffusion model for human-aware 3D scene synthesis. {\method} learns a scene diffusion model that simultaneously considers all input humans and
floor maps to generate plausible furniture layouts. During scene generation, {\method} integrates the motion collision avoidance, boundary constraint and object collision avoidance as guidance to further enhance the scene plausibility.
Additionally, we devise an automated calibration pipeline to improve the spatial accuracy and diversity of human-object interactions in existing human-aware 3D scene dataset, thereby enhancing the generation ability of {\method}. The quantitative and qualitative results showcase promising improvements in synthetic and real-world HSI datasets, demonstrating the effectiveness of our framework and calibration pipeline.

\bibliography{aaai25}


\end{document}


\maketitle




\begin{figure}[h]
\centering
\vspace{-8mm}
\includegraphics[width=\columnwidth]{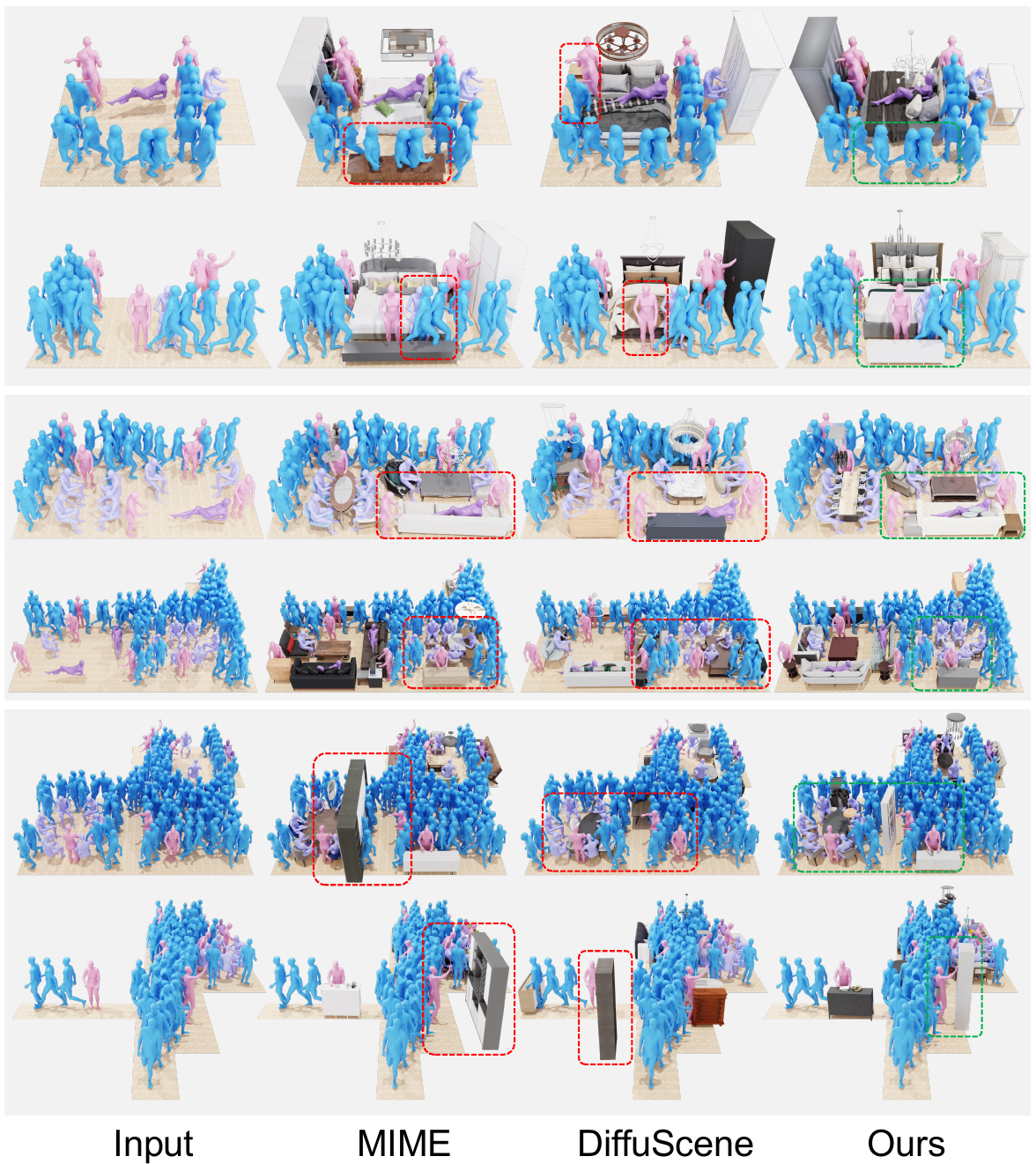}
\caption{\textbf{Qualitative comparison on the bedroom (the first two rows), living room (the second two rows) and dining room (the last two rows) in calibrated 3D FRONT HUMAN}. 
Compared with the baseline methods MIME and DiffuScene, our method generates more plausible scenes that avoid conflict with free-space humans and room boundaries, while presenting fewer overlapping objects. 
}
\vspace{-4mm}
\label{supply:compare}
\end{figure}

\begin{figure}[h]
\centering
\vspace{-8mm}
\includegraphics[width=\columnwidth]{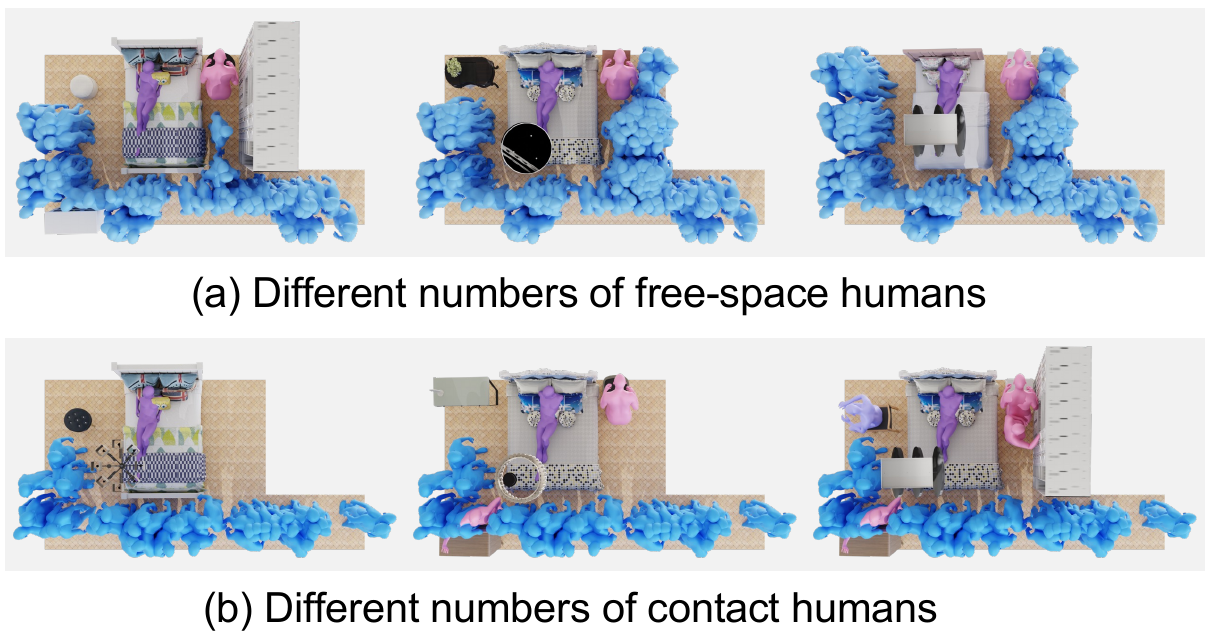}
\caption{Ablation study on varying numbers of free-space and contact humans. (a) As the number of free-space humans increases, {\method} generates fewer objects in the scene. (b) Providing more contact humans as input, {\method} generates more occupied objects to interact with them.
}
\vspace{-4mm}
\label{ablation2}
\end{figure}

\subsection{Network Details}
Our human-aware scene diffusion model is a transformer-based architecture composed of four DiT blocks with adaLN-Zero \cite{2022_scale}. Each block has a latent dimension of 512, and each attention layer consists of 8 heads. 
The timestep $t$ is embedded by a linear layer with the sinusoidal function. 
For simplicity, we encode objects and contact humans with a shared fully connected network in \cite{tang2023diffuscene}. 
To encode the free space from floor plan $\mathcal{F}$ and the free-space mask $\mathcal{FS}$, we first calculate the Hadamard product between $\mathcal{F}$ and  $\mathcal{FS}$ to obtain a 2D image mask $\mathcal{G}$, where pixel
values of 1 indicate empty regions for furniture placement and pixel values of 0 otherwise. Then, we uniformly sample 1000 points from $\mathcal{G}$ using the farthest point sampling algorithm \cite{eldar1994farthest} and transform them to the canonical space of the 3D scene. Finally, we obtain the 3D point set $\mathcal{P}\in \mathbb{R}^{1000\times3}$ as the representation of the free space within each floor plan. Next, we employ PointNet \cite{qi2017pointnet} to extract the global layout embedding from $\mathcal{P}$. 
To ensure the scene generation is aware of different contact humans and room layout, we sum the embedding vectors of each contact human, floor plan, and $t$ as conditions for each contact object, from which we regress the scale and shift parameters of each adaLN-Zero block. These parameters are then applied to modulate the embedding vector of the corresponding contact object. This strategy effectively guides the diffusion model to generate plausible contact objects for different contact humans. Note that the generation of non-contact objects only condition on the sum of the embedding vectors of layout and $t$.

\subsection{Training Details}
Following the default settings in \cite{NEURIPS2020_ddpm}, the forward process variances of our scene diffusion model are set to constants increasing linearly from $\beta_1$=$10^{-4}$ to $\beta_T$=$0.02$. Additionally, we reduce the number of diffusion steps to $T$=200 from 1000, thereby substantially speeding up the sampling process. 
During training, we sample the diffusion step $t$ from a uniform distribution at each iteration and use the Adam \cite{kingma2014adam} optimizer with a learning rate of $1\times10^{-4}$ and no weight decay. 
Following \cite{yi2022mime}, we apply random global rotation augmentation between [0, 360] degrees on the entire scene, including the floor plane, all objects, all contact humans, and the free space.
Since the number of objects in different scenes might be different, we insert the \texttt{[EMPTY]} objects into each scene as placeholders to obtain the fixed number of objects. Specifically, the maximum number $N$ of objects in the bedroom is set to 12, while the living room and dining room can accommodate a maximum of 21 objects. 
To ensure consistency with the number of objects in the scene, we expand the number of contact humans to $L$ by padding with non-contacted humans as needed. 
Finally, we train {\method} on a Nvidia Tesla A100 GPU for 625K iterations with a batch size of 128.

\subsection{Spatial-constrained Sampling Details}
The detailed algorithm of diffusion sampling with spatial-constrained guidance is provided in Algorithm \ref{algo}.

\vspace{-2mm}
\begin{algorithm}[h]
\setlength{\baselineskip}{1.3\baselineskip}
\caption{Guided sampling in {{\method}}}
\textbf{Module:} {Denoising Network $\bm{\epsilon}_\theta$, spatial-constrained guidance $\mathcal{J}=\mathcal{J}_m+\mathcal{J}_b+\mathcal{J}_o$}

\SetKwFunction{MyFunction}{sampling}
\SetKwProg{Fn}{function}{:}{}
\Fn{\MyFunction{$\mathbf{x}_{t}, C, \mathcal{J}$}}{
    
    \texttt{// Estimate clean scene}
    
    $\hat{\mathbf{x}}_0=\frac{1}{\bar{\alpha}_t}\left(\mathbf{x}_t-\sqrt{1-\bar{\alpha}_t}\bm{\epsilon}_\theta(\mathbf{x}_t, t, C)\right)$

    $\tilde{\mathbf{x}}_0 = \hat{\mathbf{x}}_0 - \gamma\nabla_{\mathbf{x}_t}\mathcal{J}(\hat{\mathbf{x}}_0, , \mathcal{FS}, \mathcal{F})$

    \texttt{// Sampling cleaner scene}

    $\mathbf{x}_{t-1} = \mathcal{N}(\mathbf{x}_{t-1};\sqrt{\bar\alpha_{t-1}}\tilde{\mathbf{x}}_0, (1-\bar\alpha_{t-1})\mathbf{I})$
    
    \KwRet{$\mathbf{x}_{t-1}$}
}

\texttt{// Spatial-constrained generation}

\KwIn{Initial scene $\mathbf{x}_T\sim \mathcal{N}(\mathbf{0}, \mathbf{I})$, contact humans $\mathcal{C}$, floor plan $\mathcal{F}$ and free-space mask $\mathcal{FS}$}  
\For{$t=T,...,1$}{
    $C=\{ \mathcal{C}, \mathcal{F}, \mathcal{FS}\}$
    
    \texttt{// Sampling with guidance}
    
    $\mathbf{x}_{t-1} = \MyFunction (\mathbf{x}_{t}, C, \mathcal{J})$
}
\Return{$\mathbf{x}_0$}\;
\label{algo}
\end{algorithm}
\vspace{-4mm}

\subsection{More Qualitative Results}
Figure \ref{supply:compare} presents more qualitative examples for different kinds of rooms. Compared with the baseline methods MIME \cite{yi2022mime} and DiffuScene \cite{tang2023diffuscene}, {\method} can generate more plausible 3D scenes that avoid conflict with free-space humans and room boundaries, while reducing collisions between objects.

\subsection{Ablation Study on Various Number of Input Humans}
In Fig. \ref{ablation2}, we further investigate the impact of input humans on scene generation by varying the number of free-space humans and contact humans provided as input.
Our qualitative results indicate that as the density of free-space humans increases, our method generates fewer objects in the scenes. Additionally, when given more contact humans, we produce more occupied objects to interact with them. 
These findings demonstrate the flexible scene generation capabilities of our method, influenced by the varying numbers of input humans.

\subsection{More Discussions}

Our method, {\method}, has demonstrated impressive results in generating plausible 3D scenes given human motion sequences and the floor plan, though it still has some limitations. 
First, we consider all generated scenes to be static, meaning our current approach does not accommodate dynamic objects involved in complex human interactions, such as moving a nightstand. This limitation restricts our method's ability to capture the full range of human-scene interactions in real-world scenarios.
In future work, it would be promising to focus on dynamic scene generation by modeling the interactions between humans and dynamic objects. 
Additionally, future research could explore human-ground interactions \cite{ma2023grammar} or incorporate contact-aware textual descriptions \cite{ma2024richcat} during scene generation, enabling our method to support a wider range of human motions, including jumping and leaping.


\bibliography{supp}